%% file: main.tex
\documentclass[letterpaper, 10 pt, conference]{ieeeconf}  %

\IEEEoverridecommandlockouts                              %

\usepackage{cite}
\usepackage{amsmath,amssymb,amsfonts}
\usepackage{algorithmic}
\usepackage{amsmath}
\usepackage{amssymb}
\usepackage{graphicx}
\usepackage{lipsum}
\usepackage{textcomp}
\usepackage{multirow}
\usepackage[utf8]{inputenc}
\usepackage{siunitx}
\usepackage{blindtext}
\usepackage{booktabs}
\usepackage{subcaption}
\usepackage{lipsum}

\usepackage{pifont} %
\usepackage{xcolor}
\def\BibTeX{{\rm B\kern-.05em{\sc i\kern-.025em b}\kern-.08em
    T\kern-.1667em\lower.7ex\hbox{E}\kern-.125emX}}
\usepackage{caption}
\usepackage{ragged2e}
\newcommand{\figuredesc}[1]{%
  \begingroup
  \par
  \justifying\small
  \noindent #1
  \par
  \endgroup}
\usepackage{braket}
\usepackage{tikz}
\usetikzlibrary{calc}
\usetikzlibrary{arrows, arrows.meta}
\usetikzlibrary{shapes,fit}
\usetikzlibrary{positioning}
\usetikzlibrary{shadings}
\usetikzlibrary{shadows.blur}

\usepackage{scalerel}
\DeclareMathOperator*{\aggop}{\scalerel*{\square}{\sum}}
\DeclareMathOperator{\BN}{BN}
\DeclareMathOperator{\SPP}{SPP}
\DeclareMathOperator{\SKPP}{SKPP}

\DeclareMathOperator{\SKPBEV}{SKPBEV}

\begin{document}

\title{\LARGE \bf Exploiting Sparsity in Automotive Radar Object Detection Networks}

\author{Marius Lippke $^{1,2}$, Maurice Quach$^{1}$, Sascha Braun$^{3}$, Daniel K\"ohler$^{3}$, Michael Ulrich$^{1}$, Bastian Bischoff$^{1}$\\and Wei Yap Tan$^{2}$%
\thanks{$^{1}$Marius Lippke, Maurice Quach, Michael Ulrich and Bastian Bischoff are with Robert Bosch GmbH, Corporate Research, Germany\newline
        {\tt\small marius.lippke@de.bosch.com}}%
\thanks{$^{2}$Marius Lippke and Wei Yap Tan are with Mannheim University of Applied Sciences, Institute of Information Technology, Germany}%
\thanks{$^{3}$ Sascha Braun and Daniel K\"ohler are with Robert Bosch GmbH, Cross-Domain Computing Solutions, Germany}%
}

\maketitle

\begin{abstract}
Having precise perception of the environment is crucial for ensuring the secure and reliable functioning of autonomous driving systems.
Radar object detection networks are one fundamental part of such systems. 
CNN-based object detectors showed good performance in this context, but they require large compute resources. 
This paper investigates sparse convolutional object detection networks, which combine powerful grid-based detection with low compute resources. 
We investigate radar specific challenges and propose sparse kernel point pillars (SKPP) and dual voxel point convolutions (DVPC) as remedies for the grid rendering and sparse backbone architectures. 
We evaluate our SKPP-DPVCN architecture on nuScenes, which outperforms the baseline by 5.89\% and the previous state of the art by 4.19\% in Car AP4.0.
Moreover, SKPP-DPVCN reduces the average scale error (ASE) by 21.41\% over the baseline.
\end{abstract}

\section{Introduction}
Having precise awareness of the environment is essential to ensure secure and dependable functioning of autonomous driving as well as driver assistance systems. 
Radar is an important sensor modality alongside cameras and lidar due to its price and weather robustness. 
Recently, object detection networks are used for this purpose \cite{Svenningsson_f63c09e038a4459dac5cb327c4869109}, also in combination with other sensors \cite{drews2022deepfusion, lei2023hvdetfusion}. 
Often, grid-based (CNN) radar object detection networks achieve a better performance than point-based methods \cite{bxu,Nscheiner,ulrich_improved_2022}. 
Research in lidar object detection strengthens this observation \cite{chen2017multi,zhou,byang,lang_pointpillars:_2019}, which can be seen as a hint for future, high-resolution radar systems. 

Grid-based object detectors typically render the radar point cloud to a 2D birds-eye-view (BEV) or 3D Cartesian grid before the features are processed using convolutional layers. 
As a result, multiple radar points fall into the same cell or voxel, while many other cells remain empty. 
Previous research \cite{ulrich_improved_2022, koehler2023} focuses on the first issue of grid rendering, while this paper investigates the latter issue of sparsity in the grid structure. 
These effects are particularly severe for sparse point clouds, such as those obtained from radar sensors, and increase with the field of view (FOV) or grid resolution. 

Sparse convolutions \cite{Deng_Shi_Li_Zhou_Zhang_Li_2021} and submanifold convolutions \cite{graham20173d} are already applied for lidar object detection networks. 
Similarly to lidar, radar point clouds have an inherently sparse structure and sparse CNNs can reduce computational complexity substantially.
In addition, sparse CNN processing improves the object detection performance, as will be shown in this paper. 

For this purpose, we propose a novel sparse grid rendering module \textit{sparse kernel point pillars} (SKPP) and a novel sparse CNN block \textit{dual point voxel convolutions} (DPVC), which can be used in the backbone of the detection network. 
SKPP extends existing grid rendering techniques, such as kernel point convolution BEV rendering (KPBEV, \cite{koehler2023}), to sparse grids (SKPBEV). 
Further, SKPP allows combining multiple feature extractors from different grid rendering techniques (multigrid rendering). 
DPVC blocks consists of two branches, one using submanifold sparse convolutions and the other using kernel point convolutions (KPConv). 
After each block, the features of the two branches are merged. 
By doing so, the network can learn from the data, which processing (submanifold convolution or KPConv) is more suitable for the representation of this particular block. 

The contributions of this paper are the following:
\begin{itemize}
    \item To the best of our knowledge, this is the first investigation of sparse CNNs for radar object detection.
    \item Novel building blocks SKPP and DVPC, which are particularly suitable for sparse radar detectors.
    \item An ablation study for the proposed components on the nuScenes benchmark. 
\end{itemize}

\begin{figure}[!t]
\centering
\includegraphics[width=1.0\linewidth]{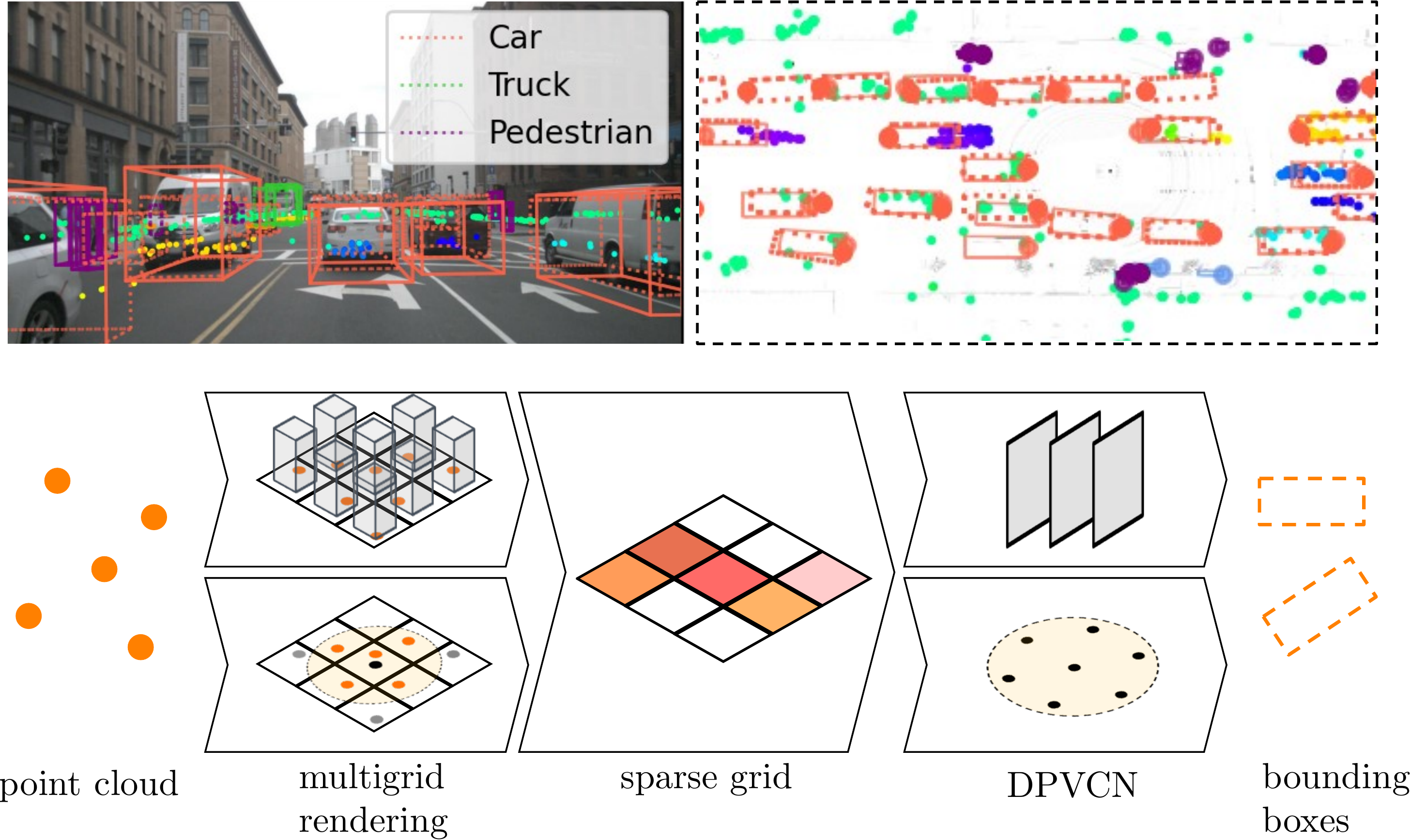}
\caption{
Object detectors based on sparse CNNs often rely on a single grid rendering method, which can lead an information loss. 
We propose the novel SKPP-DPVCN architecture for radar-only object detection. 
First, we introduce a multigrid rendering module SKPP that uses sparse KPBEV and sparse PointPillars to learn more expressive features. 
In addition, we address the limitations of submanifold sparse convolution limitations to extract local and disconnected features in sparse grids by introducing DPVCN.}
\label{fig: summary}
\end{figure}

\section{Related work}
\subsection{Radar Object Detection}

\begin{figure}
\centering
\includegraphics[width=0.7\linewidth]{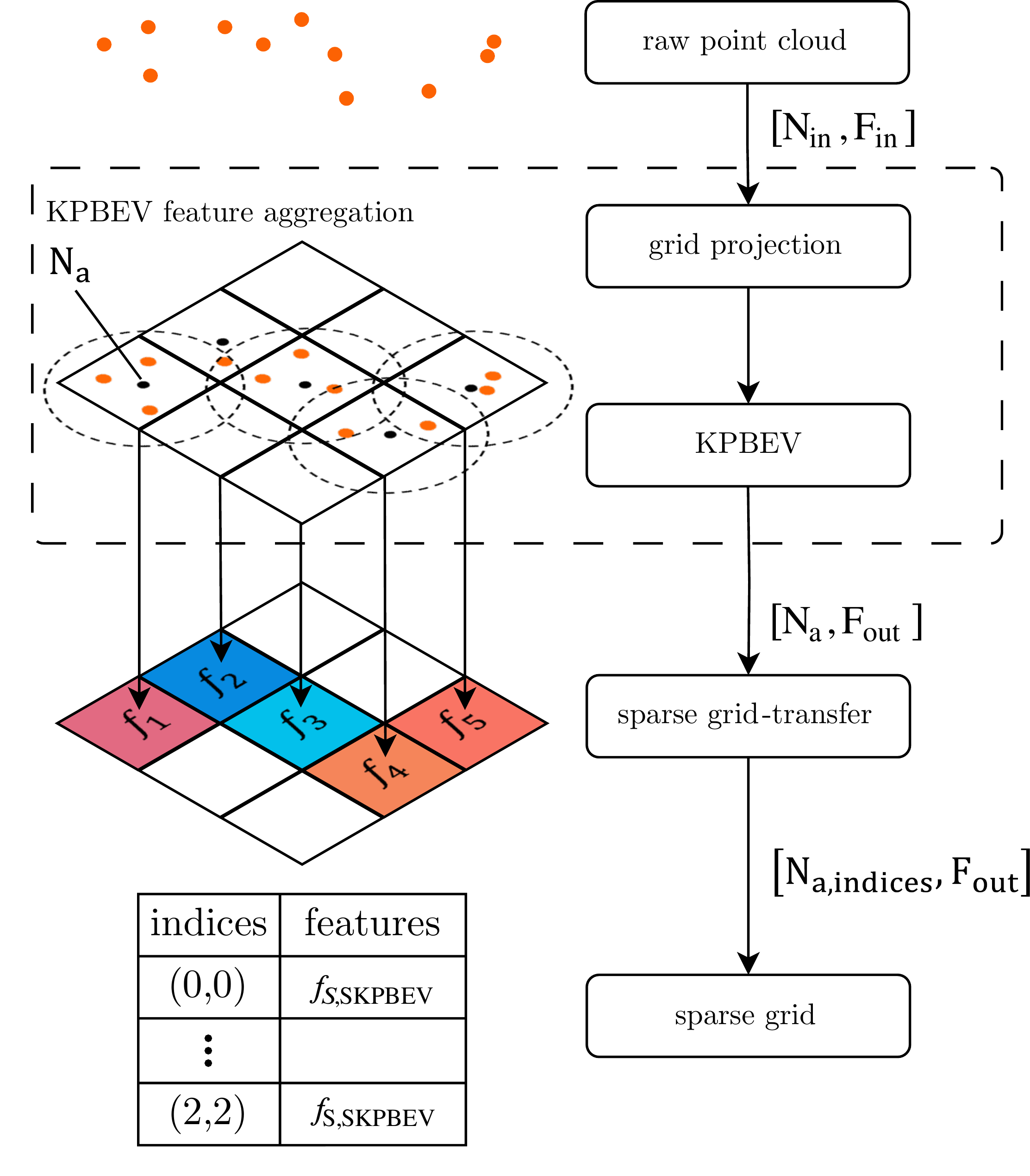}
\caption{SKPBEV extracts features at each reference point $N_a$ (centers of occupied cells) using KPConvs. The features then lie on a sparse grid.}
\label{fig: sparse_kpbev}
\end{figure}

\begin{figure*}
\centering
\includegraphics[,width=0.9\textwidth]{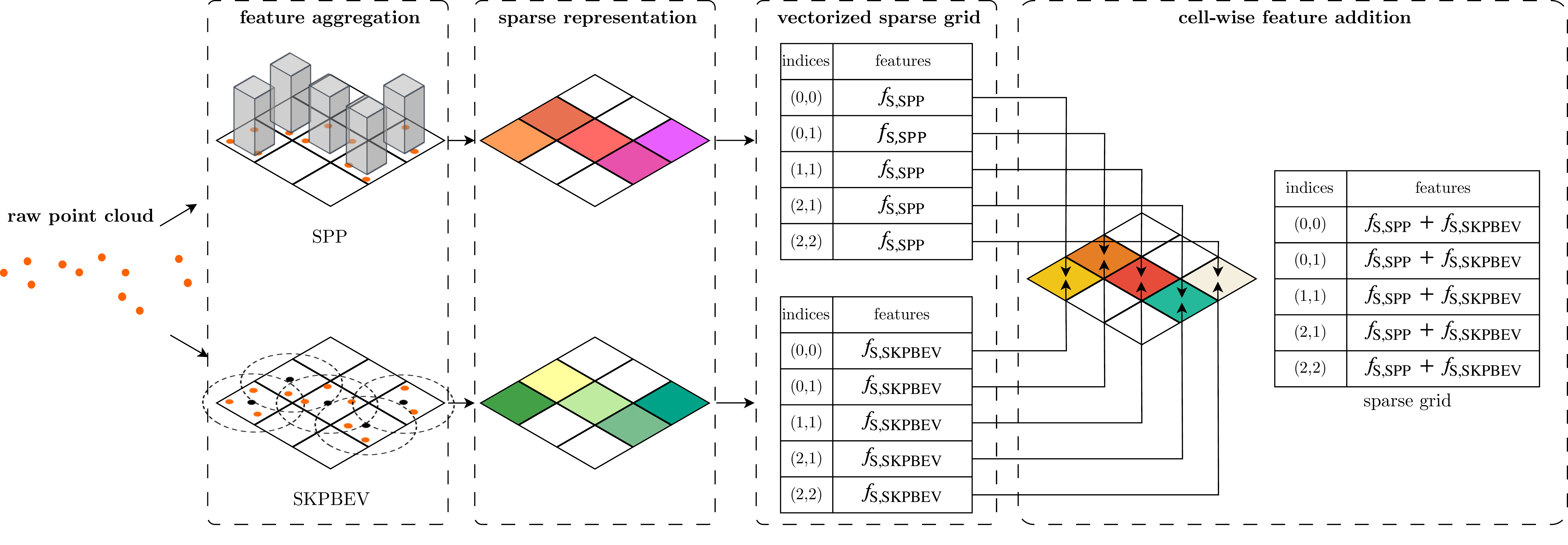}
\caption{SKPP first renders the point cloud into two sparse grids based on SPP \cite{vedder2022sparse} and the proposed SKPBEV. Finally, the features of both grids are normalized and added into a single sparse grid.}
\label{fig: SKPP}
\end{figure*}

Previous studies on radar object detection can be categorized based on the type of input data. Indeed, the research community has shown increasing interest in approaches based on radar spectra instead of radar point clouds. While 1D radar velocity spectra or 2D spectrograms (Doppler) are commonly employed for object classification \cite{ulrich2018,ulrich2018_2}, object detection networks typically rely on 3D radar spectra \cite{Palffy_2020}, incorporating range, radial velocity, and azimuth angle dimensions. These spectra are often transformed into single \cite{Brodeski_8835792} or multiple \cite{lim2019radar} 2D projections to facilitate the application of 2D CNNs. Most literature in this field utilizes CNNs due to the natural grid-based representation of spectrum data, although exceptions exist \cite{mmeyer}.

Alternatively, radar object detection can also use point clouds as input for detection networks. These approaches are typically divided into point-based and grid-based methods. Models such as PointNet \cite{qi2017pointnet} and PointNet++ \cite{qi2017pointnet_plus}, aim to directly aggregate information from point clouds. In the radar domain, these approaches have been utilized for tasks such as classification \cite{ulrich_2019_11}, semantic segmentation \cite{oschumann}, and object detection \cite{Xu9564754,Danzer_10.1109/ITSC.2019.8917000,Bansal_2020}. Recent research has also explored advanced techniques for extracting features at the point level, including graph and kernel point convolutions \cite{ulrich_improved_2022,Svenningsson_f63c09e038a4459dac5cb327c4869109,Nobis_app11062599} or transformers \cite{Bai2021}. 
On the other hand, grid-based methods transform the point cloud into a structured grid format. This enables the utilization of CNNs to extract features effectively. For this purpose, variations of the YOLO architecture \cite{Redmon_7780460} or feature pyramid networks [11], \cite{Meyer_dL3D,Lin_2017_CVPR} have been employed in studies such as \cite{Nscheiner,Dreher_2020}, to detect objects in bird's-eye-view (BEV) projections of the radar point cloud. In other works \cite{bxu,Nscheiner,btan,A_Palffy3+1D}, PointPillars \cite{lang_pointpillars:_2019} has been used to learn a more abstract grid representation of the point cloud. \cite{ulrich_improved_2022} presents a hybrid architecture that combines grid-based methods with point-based preprocessing to enhance feature learning from point clouds and improve detection performance.
\subsection{Grid Rendering of Point Clouds}
Methods for converting irregular and sparse point clouds into regular and dense grid representations play a crucial role in 3D object detection models that utilize CNNs for feature extraction. These methods can be categorized based on the resulting grid representation and the type of encoding employed to obtain cell-wise features.

For instance, MV3D \cite{chen2017multi} and PIXOR \cite{byang} utilize handcrafted features like intensity, density, and height maps to generate a BEV representation from point clouds. In contrast, VoxelNet \cite{zhou2018voxelnet} and PointPillars \cite{lang_pointpillars:_2019} leverage learnable feature extractors that apply simplified PointNets \cite{qi2017pointnet} to points within the same grid cell.
KPBEV \cite{koehler2023} on the other hand leverages the descriptive power of kernel point convolutions to encode local information during grid rendering in a multiscale manner. 
\subsection{Sparse Convolution Backbones}
In \cite{graham2014spatiallysparse,graham2015sparse}, a spatially sparse convolution technique is introduced, decreasing complexity of 3D convolutions by only considering active sites.
However, sparsity in deeper layers is lost, thus \cite{graham2017submanifold} suggests submanifold sparse convolutions (SSC), which preserve the sparsity of the output and significantly increases the computation speed. Additionally, in \cite{graham20173d}, submanifold convolutions are employed for 3D semantic segmentation.
Building upon this approach, SECOND \cite{Yan2018} introduces hierarchical encoders that progressively downsample the initial feature map using sparse convolutions before feeding it into a convolutional detection backbone. To better process sparse point clouds, \cite{vedder2022sparse} proposes sparse PointPillars to obtain a sparse grid and further process it with a sparse convolution backbone.

\section{Proposed Method}

\subsection{Overview}\label{AA}
In this paper, we introduce novel multigrid-rendering module Sparse Kernel Point Pillars (SKPP), that effectively encodes point features into a sparse grid by leveraging the strengths of both PointPillars and KPBEV methodologies, enhancing the scalability and performance of radar object detection networks. Furthermore, we propose Dual Point Voxel ConvNet (DPVCN), a hybrid grid-based backbone architecture, which efficiently processes the sparse grid by exploiting the sparse grid and point cloud duality.

\subsection{Definitions}

We define a dense 2D grid with $N_x \times N_y$ cells
\begin{equation}
   G = I \times J
\end{equation}
where $I = \{0,\dots, N_{x-1}\}$ and $J = \{0,\dots, N_{y-1}\}$.
Subsequently, we define a sparse grid $S$ where $S \subseteq G$.

To define features over a sparse grid, we adopt the notation
\begin{equation}
f_S \colon S \to \mathbb{R}^d
\end{equation}

to represent $d$-dimensional features over a sparse grid $S$.
For instance, $f(i, j)$ represents the features at occupied cell $(i, j) \in S$.
Similarly, we adopt the notation $f_G$ when operating over a dense grid $G$.
We assume in this paper that $S$ contains the occupied cells in $G$ given an input point cloud.

\subsection{Multigrid Rendering: Sparse Kernel Point Pillars (SKPP)}

Sparse grids are an efficient alternative to dense grids when processing sparse point clouds.
Typical grid projection approaches such as PointPillars \cite{lang_pointpillars:_2019} projects the point cloud onto the dense grid $G$, where each cell represents a pillar. Within each grid-cell, PointPillars aggregates the features of the points using a PointNet-based network. However, this feature aggregation process may lead to information loss, as the original spatial arrangement of the points within a cell is discarded, potentially impacting the model's ability to capture fine-grained details. 

More recent grid-rendering methods such as KPBEV \cite{koehler2023} improves information flow between points in order to learn more expressive features of local neighborhoods. 
Specifically, KPBEV includes points from adjacent cells and uses the descriptive capabilities of Kernel Point Convolution (KPConv), which enables it to better capture local context.
For each occupied grid cell, an reference point placed in the center of the grid cell.
Then, for each reference point, the neighboring points in the input point cloud are retrieved and their features are aggregated with a KPConv. 
Finally, the features are back scattered into a dense grid $G$.

However, $G$ scales linearly with the number of cells and thus scales quadratically with the distance.
In \cite{vedder2022sparse}, a sparse formulation of PointPillars (SPP) elides the feature back scattering step to $G$ and uses a sparse grid $S$ to scale lineraly with the number of occupied cells. 

The proposed SKPBEV elides the feature back scattering step of the KPBEV to $G$ as well and uses a sparse grid $S$ instead.

\begin{figure}
\centering
\includegraphics[width=1.0\linewidth]{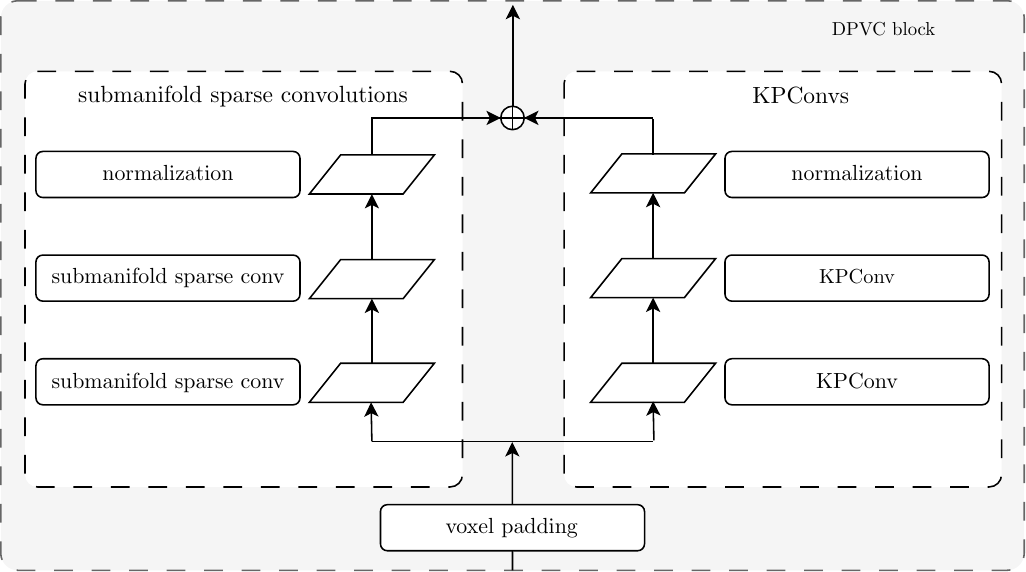}
\caption{Structure of a DPVC block. The sparse input grid is processed by a branch of submanifold sparse convolution layer (left) and another branch with KPConvs. Each convolution layer in both branches is followed by a batch normalisation and ReLU activation function. The normalized features for each branch are added, resulting in a single sparse grid as output.}
\label{fig: DPVC}
\end{figure}

\begin{figure*}
\centering
\includegraphics[width=0.8\textwidth]{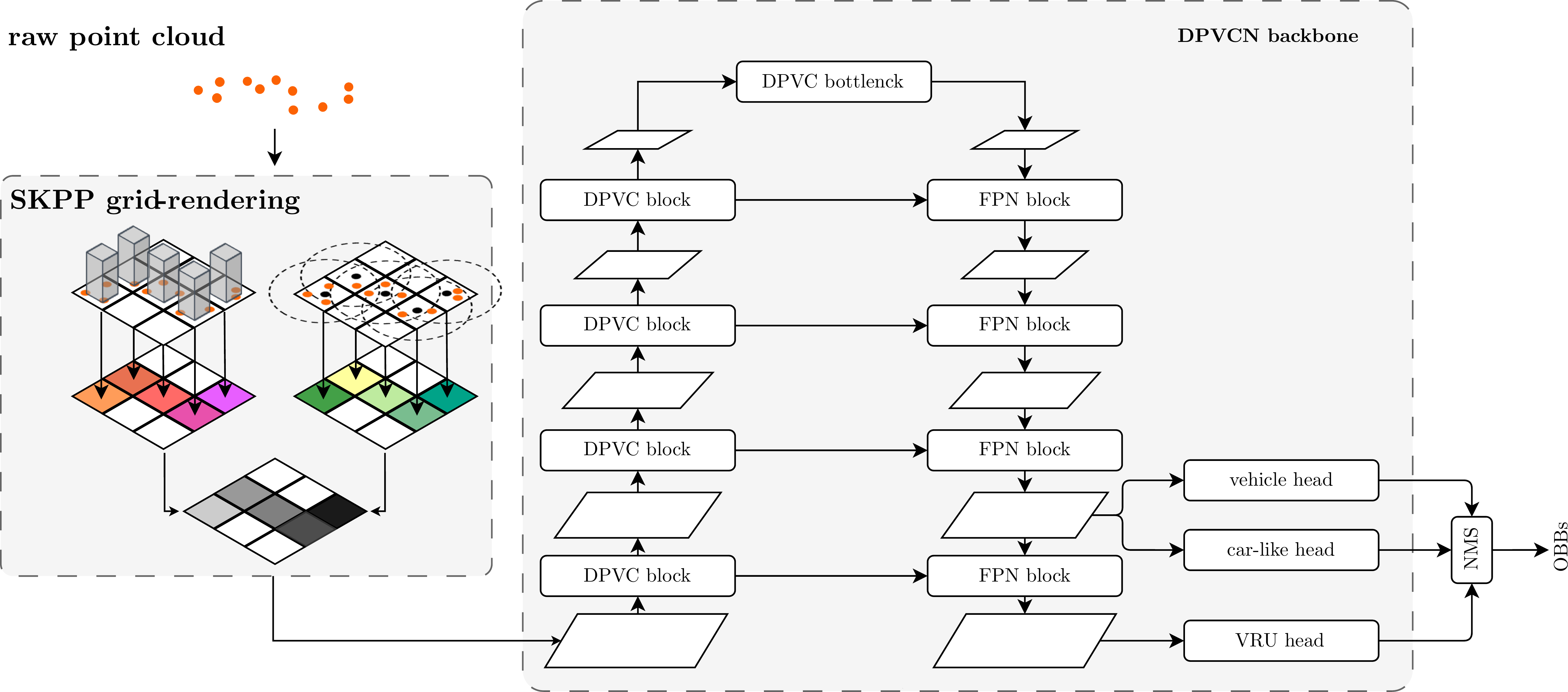}
\caption{The proposed SKPP-DPVCN first renders the input point cloud into a sparse grid $S$ with features $f_{S,\SKPP}$. Then, the features are processed by the DPVCN backbone to extract feature maps where each DPVC block downsamples by a factor of two. The FPN blocks upsample the feature map and apply a SSC block. Convolutional detection heads for different classes predict OBB proposals. Additionally, NMS is applied as a post-processing step.}
\label{fig: SSKPC_backbone}
\end{figure*}
Although the two methods SKPBEV and SPP are fundamentally different in the feature aggregation step, the features $f_{S,\SKPBEV}$ lie over the same sparse grid $S$ as $f_{S,\SPP}$.

We first propose a general multi-grid rendering framework as follows.
Let $F_{S}$ be a set of features from grid rendering methods that render a given point cloud into a corresponding sparse grid $S$.
Then, we formulate multi-grid (MG) rendering as follows
\begin{equation}
    f_{S,MG} = \aggop_{f_{S} \in F_{S}} f_{S}
\end{equation}
where $\aggop$ is an aggregation operator.
We note that typical grid rendering methods are a special case of multi-grid rendering where $|F_{S}| = 1$ and $\aggop = \sum$.

To further improve information flow from point clouds to sparse grids, we propose a novel sparse multi-grid rendering module Sparse Kernel Point Pillars (SKPP), based on the proposed SKPBEV and SPP to render both features into the same sparse grid $S$.
Specifically, SKPP is an instance of multi-grid rendering with resulting features $f_{S,\SKPP}$ where
\begin{equation}
F_{S} = \set{f_{S,\mathrm{\SPP}}, f_{S,\SKPBEV}}
\end{equation}
and
\begin{equation}
    \aggop_{f_{S} \in F_{S}} f_{S} = \sum_{f_{S} \in F_{S}} \BN(f_{S})
\end{equation}
where the aggregation first applies batch normalization (BN) before summing features.
We've found empirically that BN before summation improves performance.

\subsection{Dual Point Voxel Convolution (DPVC) Blocks}

Regular sparse convolution suffers from the dilation of all sparse features, leading to increased computational burden \cite{shi2021pvrcnn,s18103337}. Furthermore, it diminishes sparsity and blurs feature distinctions, which undermines its efficacy in distinguishing target objects from background features \cite{graham20173d}.

In contrast, submanifold sparse convolutions confine output features to the input, alleviating the computational issue but sacrificing information flow, particularly for spatially disconnected features.
While this approach is appropriate for conventional 2D convolutions operating on regularly and uniformly distributed points such as image pixels, it is less suitable for unordered sparse point clouds such as radar point clouds.

We address this problem in submanifold sparse convolution (SSC) networks by improving information flow of spatially disconnected features.
We achieve this by exploiting the dual nature of sparse grids as they can be considered simultaneously as a grid and a point cloud.
Indeed, the set of indices $(i, j) \in S$ is a sparse grid but can also be considered as a point cloud with integer coordinates.
Therefore, we propose a novel Dual Point Voxel Convolution (DPVC) block which computes submanifold sparse convolutions, which operate on sparse grids, and KPConvs, which operate on point clouds, in parallel in order to extract more representative features. First, we extend all occupied grid cells of the input sparse grid by zero-initialising their features to the eight connected non-occupied neighbouring sites. This voxel padding allows the following convolution operations to diffuse the features. Then, we consider two branches of computation within each DPVC block (Fig. \ref{fig: DPVC}). In the first branch, we start with two 3x3 SSC layers to effectively diffuse information throughout the sparse grid, followed by BN and ReLU non-linearity. Finally, a last BN is applied. In the second branch, we process the input data with two KPConv layers, each followed by BN and ReLU non-linearity. Here, a BN is applied last as well. Lastly, the results of both branches are added.

\subsection{Dual Point Voxel ConvNet (DPVCN)}
The original PointPillars backbone processes a dense pseudo-image with a feature pyramid network backbone\cite{Lin_2017_CVPR}. Our empirical findings indicate that this backbone architecture exhibits poor performance when applied to radar data, which aligns with the observations made in \cite{scheiner_object_2021}. In addition, it only processes dense grids which scale poorly with increased ranges.

Instead, we propose Dual Point Voxel ConvNet (DPVCN) (Fig. \ref{fig: SSKPC_backbone}), employing our aforementioned DPVC block for each basic block of the network.
We use a size two max-pooling operation when downsampling. We use a voxel unpooling as the opposite operation to restore the high grid resolution. We've found empirically that using the DPVC blocks only in the encoder performs best.

\begin{figure}[!htb]
\centering
\includegraphics[width=0.8\linewidth]{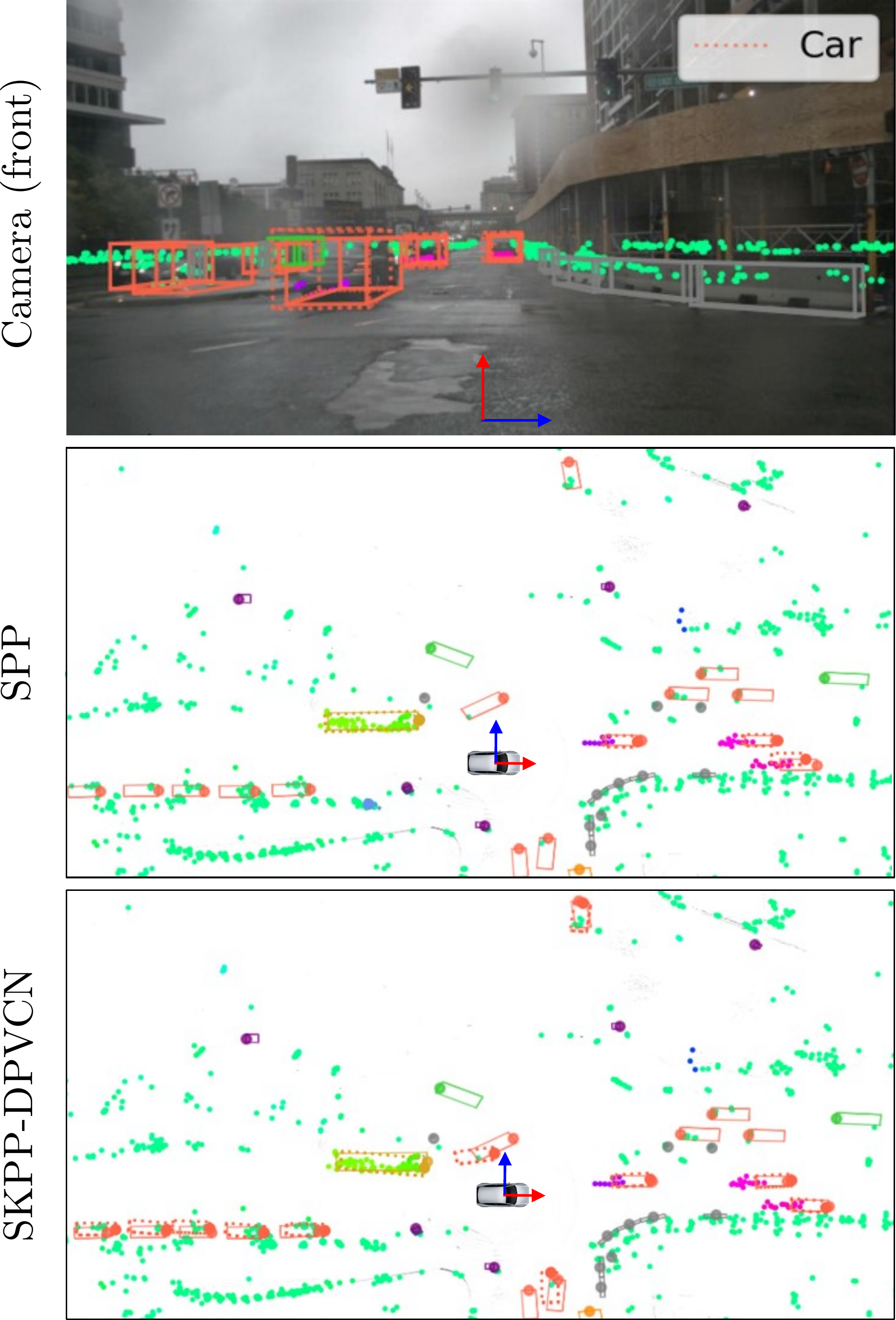}
\caption{Images of the front camera and corresponding qualitative results of SPP and SKPP-DPVCN in BEV perspective. The results show the radar point cloud as colored points (color encodes radial velocity), lidar points in grey,  ground truth boxes as solid rectangles and predicted boxes as dotted rectangles. The results illustrates the improved detection performance of SKPP-DPVCN, as both models detects cars that are missed by the SPP \cite{vedder2022sparse} baseline model.}
\label{fig: nuScenes_visual_skpppillarsCN}
\end{figure}
\subsection{SKPP-DPVCN}

Finally, the full architecture of the proposed SKPP-DPVCN (see fig. \ref{fig: SSKPC_backbone}) fuses the benefits from the proposed SKPP grid rendering and DPVCN backbone network into a single model.
The resulting model improves information flow from point cloud to the grid, scales well to farther distances with sparse grids and sparse convolutions and exhibits strong feature extraction capabilities by exploiting the duality between sparse grids and point clouds.

As the final step, detection heads process the output feature maps at different scales. We use a coarser scale for larger objects and a finer scale for small objects such as vulnerable road users (VRUs). Non-maximum suppression (NMS) is employed to refine oriented bounding boxes (OBBs) for various class categories.

\section{EXPERIMENTS}
\subsection{Experimental Setup}
\begin{table*}[ht!]
    \newcommand\modelspacing{0.154cm}
    \centering
    \caption{Ablation of the different architectures for class \textit{car} on the nuScenes test set.}
    \figuredesc{The proposed multigrid-rendering module SKPP significantly outperforms our implementation of SPP \cite{lang_pointpillars:_2019} and  SKPBEV in terms of AP4.0, mAP, ATE and ASE at a slightly lower frame rate. Our new proposed DPVCN backbone also outperforms the SSCN \cite{graham20173d} backbone in terms of AP4.0 and mAP at a lower frame rate but comes with higher computational costs. The proposed SKPP-DPVCN outperforms all other methods in terms of AP4.0, mAP, ASE and ATE.\vspace{0.25cm}} 
    \label{tab:benchmark_nuscenes}
    \input{result_table}
\end{table*}
We evaluate different tradeoffs between detection performance, computational complexity and true positive metrics with variants of SKPP-DPVCN. Specifically, we consider the dimensions of the grid rendering method (sparse PointPillars, sparse KPBEV or SKPP) and the backbone architecture whether the baseline submanifold sparse convolution (SSCN) backbone \cite{graham20173d} or our DPVCN is used. Indeed, we find in our experiments that the combination of SKPP grid rendering and DPVCN significantly improves detection performance compared to other variants. We utilize the official evaluation toolkit and metrics \cite{caesar2020nuscenes}. In the nuScenes benchmark, average precision (AP) metrics are utilized to assess the detection performance, considering various matching thresholds (0.5, 1, 2, and 4 meters) between the ground truth and predictions. For our evaluation we focus on the average precision for a matching threshold of 4 m (AP4.0). Furthermore, Average Scale Error (ASE) quantifies the intersection over union (IOU) after aligning orientation and translation ($1 - IoU$) and Average Orientation Error (AOE) quantifies the smalles yaw angle difference between pr8ediction and ground truth.
Similar to previsous research \cite{ulrich_improved_2022,9455172}, we focus on the AP4.0 for the class car because other objects are hardly distinguishable using only the low resolution radar in the nuScenes dataset. Consequently, numerous objects either lack reflections entirely or exhibit only a minimal number of reflections. Moreover, the sparsity of and the absence of elevation in nuScenes' radar data hinder the accurate classification of certain objects. This limitation results in confusion when distinguishing between classes like buses and trucks.
\subsection{Models}
We evaluate the performance of the proposed SKPP-DPVCN method and perform an ablation study to quantify the impact of each individual contribution. To this end, we consider four different evaluation settings:
\begin{itemize}
    \item \textbf{SPP (baseline)}: Our re-implementation from \cite{lang_pointpillars:_2019} which utilizes the voxelization feature encoder of PointPillars to convert the point cloud into a vectorized sparse BEV feature map which is processed by a SSCN backbone network.
     \item \textbf{SKPBEV (ours)}: A sparse version of the KPBEV feature encoder with a SSCN backbone network.
    \item \textbf{SKPP (ours)}: A multigrid-rendering method based on SPP and SKPBEV to render the features into a sparse BEV feature map which is processed by a SSCN backbone network.
    \item \textbf{SPP-DPVCN (ours)}: The baseline sparse PointPillars with our DPVCN backbone network.
    \item \textbf{SKPBEV-DPVCN (ours)}: Our SKPBEV in combination with our DPVCN backbone
    \item \textbf{SKPP-DPVCN (ours)}: Our high performance setting which combines the SKPP multigrid-rendering module with our DPVCN backbone network.
\end{itemize}
\subsection{Experiment parameters}
The input to the network is a radar point cloud that is aggregated over seven consecutive measurements. Each point in the cloud contains the 2D Cartesian coordinates $x,y$, the ego-motion compensated radial velocity $v_r$ and the RCS. We perform object detection on a grid ranging from -60 to 60 meters in both the x- and y-direction with an initial cell size of $s_0 = 0.5\ m$. In order to process the input point cloud, all layers prior to the detection backbone utilize $F_{out} = 32$ channels. Whenever we use SKPBEV, we use 15 kernel points with a convolution radius of $1.5\ m$ and whenever we use KPConv in the DPVCN, we use a convolution radius of $3.75\ m$. For the detection backbones SSCN and DPVCN, we use $F_{out,encoder}=\{72,96,128,146,160\}$ channels for the encoder blocks. Additionally, we augment the RCS by adding a value $\Delta RCS$ from a normal distribution with zero mean and a standard deviation of $\sigma = 0.7$  to each reflection point. For each model variant, we conduct 5 trials and compute the average of metrics to accommodate for training stochasticity. Each model is trained for 30 epochs using a batch size of 32.
\section{Results}

Tab. \ref{tab:benchmark_nuscenes} shows the quantitative results of the different methods on the nuScenes validation set. We see that our SKPP-DPVCN has a relative AP4.0 improvement of 5.89\% and an absolute improvement of 2.53\% compared to the SPP-like \cite{vedder2022sparse} baseline module. Simultaneously, we see that in relative terms the ASE improves by 21.41\% and the AOE by 5.73\% relatively to the baseline. Furthermore, we find that the framerate deteriorates from 61.92 Hz to 31.48 Hz.
Tab. \ref{tab:benchmark_sota} shows that SKPP-DPVCN also outperforms the dense state-of-the-art radar object detection model KPPillarsBEV \cite{koehler2023} by 4.19\% AP4.0. 
As an ablation study, we analyse the effects of both the SKPP multigrid rendering module and the DPVCN backbone separately. Comparing the SKPP module with the SPP \cite{vedder2022sparse}, it achieves an absolute AP4.0 increase of 0.73\%. At the same time, the frame rate hardly deteriorates (Table \ref{tab:benchmark_nuscenes}) and both ASE and AOE improve by 3.64\% and 2.08\%, respectively. This performance improvement comes at no additional cost. SKPP benefits from the inherent advantages of both grid rendering methods. 
\begin{table}
    \newcommand\modelspacing{0.154cm}
    \centering
    \caption{Benchmark of the different architectures for class \textit{car} on the nuScenes test set.}
    \figuredesc{The proposed SKPP-DPVCN significantly outperforms the dense methods \cite{ulrich_improved_2022,koehler2023} in terms of AP4.0, ASE and AOE. \vspace{0.25cm}} 
    \label{tab:benchmark_sota}
    \input{benchmark_table}
\end{table}
The new DPVCN backbone with SPP-like \cite{vedder2022sparse} grid-rendering achieves an absolute improvement of AP4.0 of 1.49\% compared to the baseline SSCN \cite{graham20173d}. Additionally, a relative improvement of ASE and AOE by 7.47\% and 1.48\% respectively is obtained. However, the frame rate decreases to 31.48 Hz.
The trade-offs between detection performance and computational complexity are shown in Tab. \ref{tab:benchmark_nuscenes}. In summary, both SKPP and DPVCN contribute to the performance of radar object detection networks. The modularity of our proposed SKPP-DPVCN helps in designing suitable architectures that balance detection performance and interference time. When maximum detection performance is desired, the proposed multiscale SKPP-DPVCN architecture is the best choice and achieves an AP4.0 of 45.51\% in our experiments. If an increase in compute resources is undesirable, the SKPP multigrid rendering module significantly increases detection performance with negligible increase in interference time. 

Qualitative results of the high detection performance of this method can be found in Fig. \ref{fig: nuScenes_visual_skpppillarsCN}, showing the results from SKPP and SKPP-DPVCN in comparison to the SPP baseline based on a scene in the nuScenes validation set. In a scenario with limited computational resources, SKPP may be advantageous as it achieves significantly better detection performance at the same inference speed. Additionally, a further increase in detection performance is seen with SKPP-DPVCN, although this comes with a higher interference speed. 
\section{Conclusion}
In conclusion, we have shed light on the limitations of grid-based approaches and highlighted the importance of preserving essential information of the point cloud data. By introducing the novel multigrid-rendering method SKPP, we have improve detection performance and have overcome the tradeoffs faced by previous models.

Furthermore, our exploration of the duality between sparse grids and point clouds has led to the development of the DPVCN backbone architecture. This innovative approach leverages the descriptive power of KPConvs and SSCs to effectively extract information from spatially disconnected features, addressing the challenge posed by unstructured sparse radar data.
We have proposed SKPP-DPVCN which outperforms the current state-of-the-art \cite{koehler2023} for radar object detection on nuScenes and surpassing the baseline model. We have also performed an ablation study showing the individual contributions of SKPP and DPVCN. 
\bibliographystyle{IEEEtran}
\bibliography{main}

\end{document}

%% file: result_table.tex
        
\begin{tabular}{@{}llccccc@{}}
	\toprule
	\textbf{Grid-Rendering} & \textbf{Backbone} & \textbf{AP4.0 Car [\%] $\uparrow$} &\textbf{mAP Car}[\%] $\uparrow$ &  \textbf{ASE [m] $\downarrow$} & \textbf{AOE [deg.] $\downarrow$}& \textbf{FPS [Hz] $\uparrow$}\\ \midrule
	\multirow{1}{*}{SPP-like \cite{vedder2022sparse} (baseline)}  &SSCN-like \cite{graham20173d} (baseline)& 42.98 & 24.29   & 0.495& 39.26&\textbf{61.92}\\
	\multirow{1}{*}{SKPBEV (ours)} & SSCN-like\cite{graham20173d} (baseline)& 43.10&24.39  & 0.509 & 39.47& 60.16 \\
	\multirow{1}{*}{SKPP (ours)} & SSCN-like \cite{graham20173d} (baseline)&  43.71&24.90   & 0.477& 38.44& 57.97\\
        \multirow{1}{*}{SPP-like \cite{vedder2022sparse} (baseline)} & DPVCN (ours)&44.47 &25.20 & 0.458& 38.68& 31.98\\
	\multirow{1}{*}{SKPBEV}  & DPVCN (ours)& 44.78&25.32  &0.460 & 38.04& 30.21\\
        \multirow{1}{*}{SKPP (ours)}  &DPVCN (ours)& \textbf{45.51} & \textbf{25.98} & \textbf{0.389} & \textbf{37.01}& 31.48\\
\end{tabular}

%% file: benchmark_table.tex
\begin{tabular}{@{}lccc@{}}
	\toprule
	\textbf{Method} & \textbf{AP4.0 Car [\%]$\uparrow$} &  \textbf{ASE [m]$\downarrow$} & \textbf{AOE [deg.]$\downarrow$}\\ \midrule
        \multirow{1}{*}{PointPillars \cite{lang_pointpillars:_2019}} & 38.31   & 0.48 & 38.48\\
        \multirow{1}{*}{KPPillars \cite{ulrich_improved_2022}} & 40.80   & 0.5 & 37.87\\
        \multirow{1}{*}{KPBEV \cite{koehler2023}}  &41.21 & 0.49 & 38.49\\
        \multirow{1}{*}{KPPillarsBEV \cite{koehler2023}}  &43.68 & 0.44 & 37.31\\
        \multirow{1}{*}{SKPP (ours)}  &43.71 & 0.48 & 38.44\\
        \multirow{1}{*}{SKPP-DPVCN (ours)}  & \textbf{45.51} & \textbf{0.39} & \textbf{37.01}
\end{tabular}